\documentclass[journal]{IEEEtran}
\usepackage{authblk}
\usepackage{amsmath}
\usepackage{amssymb}
\usepackage[linesnumbered,ruled,vlined]{algorithm2e}

\ifCLASSINFOpdf
\else
\fi

\usepackage{amsmath}
\usepackage{graphicx}
\usepackage{xspace}
\usepackage{xcolor}
\usepackage[utf8]{inputenc}
\usepackage{booktabs}
\usepackage{comment}
\usepackage{algorithmic}
\begin{document}
\title{Data-Driven Antenna Miniaturization: A Knowledge-Based System Integrating Quantum PSO and Predictive Machine Learning Models }
\author{Khan Masood Parvez, Student Member, IEEE, Sk Md Abidar Rahaman, and Ali Shiri Sichani, Member, IEEE% 
\thanks{K. M. Parvez is with the Department of Electronics and Communication Engineering, Aliah University, Kolkata, India. Email: masood.ece.rs@aliah.ac.in.}%
\thanks{S. M. A. Rahaman is with the Department of Computer Science, Tarakeswar Degree College, University of Burdwan, West Bengal, India. Email: abidar.ce@gmail.com.}%
\thanks{A. S. Sichani is with the Department of Electrical Engineering and Computer Science, University of Missouri, Columbia, Missouri, USA. Email: asp9f@missouri.edu.}%
}%
%
%\maketitle

\markboth{Journal of \LaTeX\ Class Files,~Vol.~XX, No.~XX, XX~20XX}%
{Shell \MakeLowercase{\textit{et al.}}: Bare Demo of IEEEtran.cls for IEEE Journals}



\maketitle

\begin{abstract}
The rapid evolution of wireless technologies necessitates automated design frameworks to address antenna miniaturization and performance optimization within constrained development cycles. This study demonstrates a machine learning-enhanced workflow integrating Quantum-Behaved Dynamic Particle Swarm Optimization (QDPSO) with ANSYS HFSS simulations to accelerate antenna design. The QDPSO algorithm autonomously optimized loop dimensions in 11.53 seconds, achieving a resonance frequency of 1.4208 GHz – a 12.7 percent reduction compared to conventional 1.60 GHz designs. Machine learning models (SVM, Random Forest, XGBoost, and Stacked ensembles) predicted resonance frequencies in 0.75 seconds using 936 simulation datasets, with stacked models showing superior training accuracy (R²=0.9825) and SVM demonstrating optimal validation performance (R²=0.7197). The complete design cycle, encompassing optimization, prediction, and ANSYS validation, required 12.42 minutes on standard desktop hardware (Intel i5-8500, 16GB RAM), contrasting sharply with the 50-hour benchmark of PSADEA-based approaches. This 240× acceleration eliminates traditional trial-and-error methods that often extend beyond seven expert-led days. The system enables precise specifications of performance targets with automated generation of fabrication-ready parameters, particularly benefiting compact consumer devices requiring rapid frequency tuning. By bridging AI-driven optimization with CAD validation, this framework reduces engineering workloads while ensuring production-ready designs, establishing a scalable paradigm for next-generation RF systems in 6G and IoT applications.
\end{abstract}


% \fi
%\IEEEpeerreviewmaketitle

\section{Introduction}
The antenna [1] serves as a crucial element in any consumer wireless communication system, often occupying the most substantial physical volume within these configurations. As contemporary consumer electronic devices increasingly trend towards smaller and more compact designs, the imperative for antennas to adapt accordingly and integrate seamlessly into these systems has become even more significant. Consequently, the advancement of antenna miniaturization techniques [2] has become a vital necessity. Antenna miniaturization, though longstanding, remains challenging, with slot antennas using Co-Planar Waveguide ( CPW ) or microstrip lines widely studied for their compactness and adaptability. Over the years, notable progress has been achieved in miniaturization techniques in the literature [3], [4], [5], [6], [7]. Among these developments, the integration of an inductive or capacitive structure within slot antennas, along with a microstrip-fed line, has been thoroughly explored in the existing literature [3], [4]. Furthermore, CPW-fed slot antennas have experienced considerable miniaturization through innovative methods such as loop loading [5] and the application of slit and strip loading techniques [6], both of which are well documented in [7]. A miniaturized wideband circularly polarized (CP) antenna using a hybrid metasurface structure has been proposed in [8]. Capacitive slots help reduce size while maintaining performance in [9]. High-gain bow-tie antennas [10], 3D corrugated ground structures [11], and dispersive materials [12] further enhance miniaturization. Reactive impedance and polarization conversion metasurfaces improve bandwidth and reduce RCS [13]. Miniaturized antennas are widely used in wireless and biomedical applications [14], [15].
\begin{comment}
A miniaturized wide band circularly polarized (CP) antenna has been proposed using a hybrid metasurface structure [8]. Capacitive slots are crucial for minimizing antenna size while maintaining high performance [9], with a miniaturized high-gain bow-tie antenna design presented in [10], and an innovative 3-D corrugated ground structure for microstrip antenna miniaturization introduced in [11]. A study of compact wideband slot antennas [12] explored the use of specialized dispersive materials. The combination of reactive impedance surfaces with polarization conversion metasurfaces [13] has been explored to address broadband radar cross-section (RCS) reduction, antenna miniaturization, and bandwidth enhancement. In addition to their integration into consumer wireless devices, miniaturized antennas have been extensively deployed in a broad spectrum of biomedical applications [14], [15].
\end{comment}
The integration of artificial intelligence (AI) into antenna design and propagation has garnered significant attention in recent years, revolutionizing traditional methodologies and improving performance metrics. Deep learning techniques are reshaping the field of antennas and propagation, as highlighted in recent research [16], [17].
\begin{comment}
Further studies explore the use of machine learning algorithms to analyze large-scale unstructured data in this domain. These advancements are contributing to significant progress in antenna design and electromagnetic wave propagation analysis.The study in [18] introduced particle swarm optimization (PSO) for electromagnetic applications, illustrating its effectiveness in horn antenna design. Building on this, [19] enhanced the approach with quantum PSO, emphasizing the flexibility and capability of swarm intelligence in achieving optimized design solutions.A hybrid smart quantum particle swarm optimization method, tailored for multimodal electromagnetic design problems, is described in [20].
\end{comment}
Machine learning algorithms are advancing antenna design and wave propagation analysis by handling large-scale unstructured data. PSO has shown effectiveness in horn antenna design [18], with quantum PSO enhancing flexibility and optimization [19].
Recent advances in antenna design leverage multi-modal multi-objective particle swarm optimization (PSO) with self-adjusting strategies for diverse challenges [21]. Hybrid methods combining traditional optimization with machine learning, such as Taguchi binary PSO for antenna design [22], [23], have shown effectiveness. AI-driven approaches have enhanced antenna geometry design [24], optimization frameworks [26], and thin-wire antenna design using genetic algorithms [27]. Self-adaptive surrogate models [28] and trust-region Bayesian optimization for simulation-driven designs [29] further improve efficiency. Response feature surrogates enhance tolerance optimization [30], [31], with generalized formulations improving input characteristic optimization [32]. Machine learning has optimized double T-shaped monopole designs [33], extended to 3D printed dielectrics [34]. Deep learning improves beamforming and millimeter-wave antenna design [35], [36].
Researchers have introduced various advanced optimization techniques for antenna design. These include a multilayer machine-learning approach for robust antenna design [37], evolutionary algorithms specifically tailored for antenna arrays [38], and genetic algorithm-based branching methods for antennas [39], [40], [41]. Additionally, an intelligent antenna synthesis method has been proposed [42]. These approaches represent significant advancements in the field of computational antenna design and optimization.

The literature reflects a shift toward AI and machine learning in antenna design, enhancing performance across applications. This review explores innovative AI-driven approaches for automated antenna design, where users specify requirements, and the system generates design parameters and fabrication instructions for the machines to construct prototypes.These prototypes could be rapidly integrated into consumer electronics and a wide range of industrial applications, substantially expediting the fulfillment of wireless market demands within an exceptionally short period.

\begin{figure}
    \centering
    \includegraphics[width=0.8\linewidth]{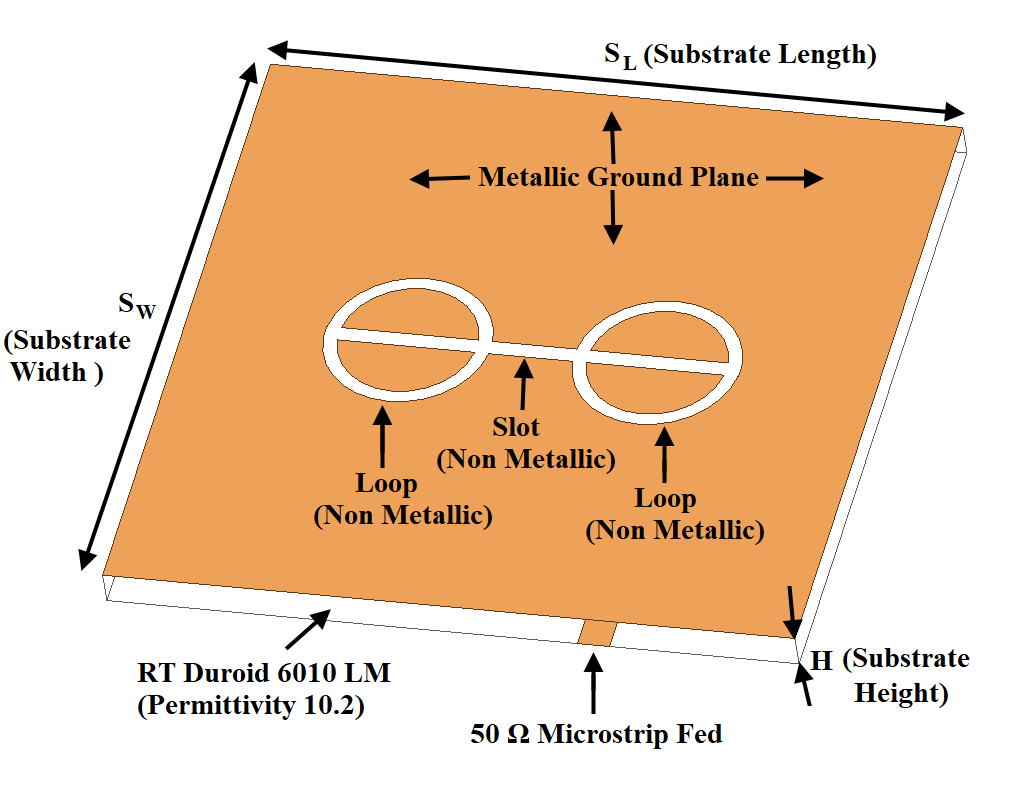} % Reduced size
    \caption{3D view of Slot antenna loaded with two loops.}
    \label{fig:enter-label}
\end{figure}

\begin{comment}
The design of a slot antenna as a case study is explored that incorporates loops at both ends of the slot to achieve a substantial reduction in resonance frequency. The integration of these loops is strategically implemented to enhance miniaturization while preserving optimal performance. To determine the precise dimensions of the loops that maximize frequency reduction, we employ Quantum-Behaved Dynamic Particle Swarm Optimization (QDPSO). This method allows us to fine-tune the loop dimensions, ensuring enhanced performance and compactness. The combination of computational optimization techniques with machine learning-based predictive modeling establishes a systematic and data-driven framework for designing compact and high-performance slot antennas. 
\end{comment}
A slot antenna design with loops at both ends reduces resonance frequency and enhances miniaturization while maintaining performance. Quantum-Behaved Dynamic Particle Swarm Optimization (QDPSO) is used to optimize loop dimensions, ensuring improved performance and compactness. This data-driven approach combines computational optimization with machine learning for efficient slot antenna design. A data set of 936 simulation results generated using ANSYS HFSS [43] serves as the foundation for training machine learning algorithms, including SVM [44], random forest [45], XGBoost [46], and Stacked Model [47] allowing precise resonance frequency predictions based on loop dimension variations. The method efficiently tackles miniaturization and frequency tuning, providing a scalable solution for advanced wireless systems in record time.
The paper is organized as follows: Section II covers antenna miniaturization, Section III discusses quantum-behaved dynamic PSO, Section IV addresses dataset integration with ML, Section V analyzes frequency prediction, Section VI highlights design automation, and Section VII summarizes key findings.

\section{Antenna design With Miniaturization Technique}
\begin{comment}
The slot antenna is an highly compact, low-profile radiating structure, making it highly suitable for wireless applications, particularly in consumer electronics. Due to their space-efficient design, slot antennas are seamlessly integrated into smartphones, tablets, and laptops, enabling essential functionalities such as Wi-Fi, Bluetooth, and cellular communication. For efficient radiation, a slot antenna must have a length equivalent to half a wavelength. The excitation through a voltage source ensures that the voltage across the aperture at the ends of the slot is zero because of the presence of shorting posts on both sides. As a result, the zero voltage condition at the edges of the slot leads to a maximum voltage amplitude occurring at a point located one quarter wavelength from each edge, precisely at the center of the slot. Loops were incorporated at both ends of the slot to preserve the maximum voltage amplitude in the center, ensuring that the radiation properties of the slot antenna remain unchanged. 
\end{comment}

The slot antenna is a compact, low-profile radiating structure ideal for wireless applications in consumer electronics like smartphones, tablets, and laptops, supporting Wi-Fi, Bluetooth, and cellular communication. For efficient radiation, the slot length should be half a wavelength. Excitation through a voltage source creates zero voltage at the slot edges due to shorting posts, resulting in maximum voltage at the center, one-quarter wavelength from each edge. Loops at both ends maintain this central maximum voltage, preserving the radiation properties of antenna.The three-dimensional schematic of the proposed antenna design is shown in Fig. 1.
Miniaturization can be conceptualized through two distinct approaches. Since the wavelength of an antenna is inversely related to the resonant frequency, one method involves reducing the physical dimensions of the antenna while maintaining a constant resonant frequency. Conversely, the alternative strategy focuses on preserving the physical size of the antenna while lowering the resonant frequency. This study employs the latter approach to achieve miniaturization.
\begin{comment}
Miniaturization can be conceptualized through two distinct approaches. It is a well-established fact that the wavelength of an antenna is inversely proportional to the resonant frequency of the antenna. The first approach involves maintaining a fixed resonant frequency while reducing the physical dimensions of the antenna. Alternatively, the second approach focuses on preserving the physical dimensions of the antenna and lowering the resonant frequency [40]. In this study, the latter approach has been adopted to achieve antenna miniaturization.
\end{comment}
\begin{figure}
    \centering
    \includegraphics[width=0.7\linewidth]{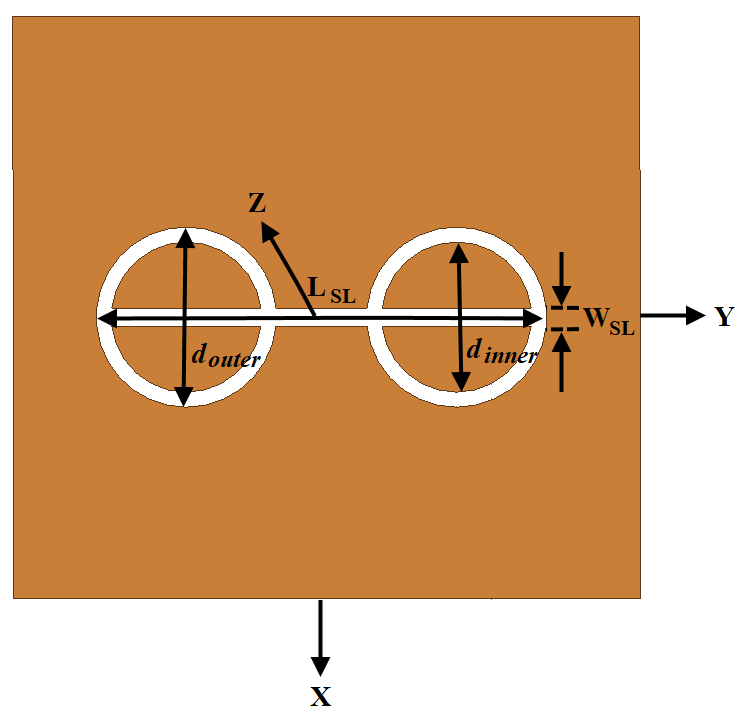}
    \caption{Top surface of slot antenna with loop.}
    
    \includegraphics[width=0.7\linewidth]{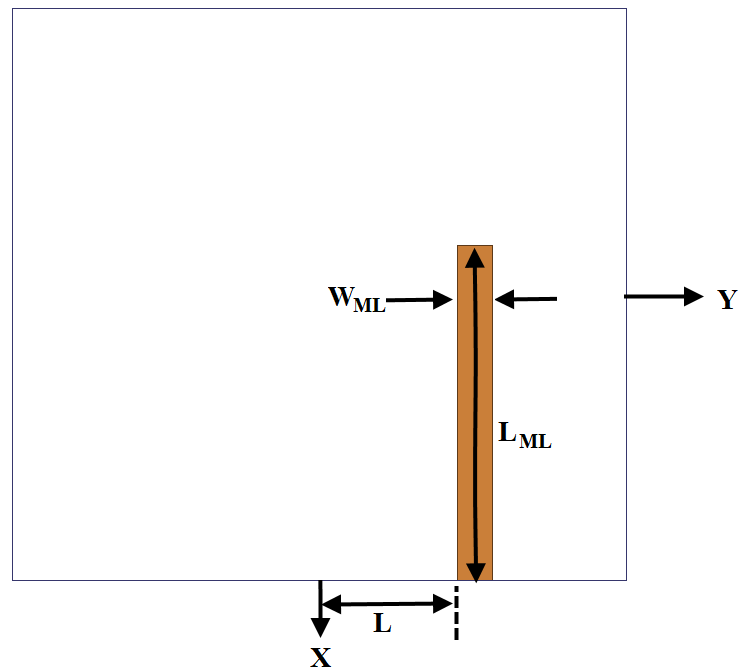}
    \caption{Bottom surface of slot antenna with loop.}
    \label{fig:enter-label}
\end{figure}
%\vspace{1cm}
\begin{table}[h]
    \centering
    \caption{Parameter Specifications for Antenna Design}
    \begin{tabular}{@{}lll@{}}
        \toprule
        \textbf{Category} & \textbf{Parameter} & \textbf{Value} \\ 
        \midrule
        Substrate Dimensions & Length & $L_S = 100.00 \, \text{mm}$ \\ 
                                & Width  & $W_S = 100.00 \, \text{mm}$ \\ 
                                & Height  & $H = 2.54 \, \text{mm}$ \\ 
        \midrule
        Slot Dimensions      & Length & $L_{SL} = 30.00 \, \text{mm}$ \\ 
                                & Width  & $W_{SL} = 1.20 \, \text{mm}$ \\ 
        \midrule
        Microstrip Feed Line      & Length       & $L_{ML} = 53.60 \, \text{mm}$ \\ 
                                & Trace Width  & $W_{ML} = 2.30 \, \text{mm}$ \\ 
                                & Stub Length   & $L_{STB} = 3.00 \, \text{mm}$ \\ 
                                & Offset        & $L = 9.00 \, \text{mm}$ (y-axis) \\ 
        \bottomrule
    \end{tabular}
\end{table}
\par
\begin{comment}
The loops have been strategically placed at the ends of the slot to allow miniaturization of the antenna. The loops located at both ends of the slot [5]-[7] demonstrate symmetrical geometry and equivalent electromagnetic properties. 
\end{comment}
Loops [5]–[7] are strategically positioned at both ends of the slot to enable antenna miniaturization, ensuring symmetrical geometry and consistent electromagnetic properties.
In a half-wavelength slot, capacitive effects dominate below the resonance frequency, requiring compensation to achieve a lower resonance frequency.The loops positioned at either end of the slot exhibit an inductive nature, effectively compensating for the capacitive reactance present in the slot. This compensation creates the possibility of achieving a lower resonance frequency. The inner ($d_{\text{inner}} $) and outer ($ d_{\text{outer}} $) diameters  of the loop are critical in determining its width, a parameter that significantly influences the inductive effect. This inductive phenomenon is directly associated with the reduction in resonance frequency relative to the reference slot [7] frequency of 2.27 GHz. Therefore, the objective is to determine the optimal loop width to achieve the maximum reduction in resonance frequency. Fig. 2 illustrates the top surface of the antenna design to clarify the concept of integration between the slot and the loops. It is important to note that, for the slot and loops, the metallic portions are removed at the locations of the slot and loop on the ground plane.
A $50  \Omega$ microstrip feed line excites the slot antenna, positioned as a strip conductor on the bottom surface of the dielectric substrate, as shown in Fig. 3. The dimensions of the substrate, slot, and microstrip line are provided in Table I. The characteristics of the low-loss dielectric substrate RT Duroid 6010LM, with a relative permittivity $( \epsilon_r ) $ of 10.2, were selected for use in the simulations.

\section{ Quantum-Behaved Dynamic Particle Swarm Optimization}
Quantum-Behaved Dynamic Particle Swarm Optimization (QPSO) [19] combines quantum mechanics with traditional PSO to enhance search efficiency. Unlike classical PSO, which follows Newtonian dynamics, QPSO uses quantum-based probabilistic movement, improving exploration and exploitation while effectively avoiding local optima and achieving faster convergence toward global solutions.
\begin{comment}
Quantum-Behaived Dynamic Particle Swarm Optimization [19] is an advanced optimization technique that integrates principles of quantum mechanics with the traditional PSO algorithm. Unlike classical PSO, which relies on Newtonian dynamics, QPSO utilizes quantum behaviors to enhance the exploration and exploitation capabilities of the search process. This method allows particles to move in a probabilistic manner, enabling them to escape local optima and converge more effectively toward global solutions.
\end{comment}
The OptiSLang, developed by ANSYS [43], serves as an advanced platform for process integration and robust design optimization across diverse engineering domains. For antenna design, OptiSLang employs a diverse array of cutting-edge optimization methodologies, including Particle Swarm Optimization [20], [21], [23], Surrogate-Based Optimization [28] and Genetic Algorithms [39], [40], [41]. After careful consideration, we opted QDPSO to refine the antenna configuration, based on the following compelling rationales: \textbf{1-Simplified Tuning:} QDPSO requires only one control parameter, simplifying implementation compared to OptiSLang's multi-parameter setup.\textbf{2-Faster Convergence:} The quantum-inspired search in QDPSO accelerates convergence, identifying optimal antenna designs more quickly than OptiSLang.
\textbf{3-Enhanced Exploration:} QDPSO’s quantum behavior improves solution space exploration, effectively avoiding local minima.
\textbf{4- for Nonlinear Problems:} QDPSO excels in optimizing complex, nonlinear antenna designs, outperforming some OptiSLang algorithms.
\textbf{5-Tailored for Electromagnetics:} QDPSO is designed for electromagnetic problems, making it more effective for antenna design than the general-purpose OptiSLang.
\textbf{6-Lower Computational Cost:} QDPSO’s efficient search and minimal parameters reduce computational demands compared to OptiSLang’s complex strategies.
\begin{comment}
\textbf{1- Simplified Parameter Tuning:} QDPSO requires only one control parameter, making it easier to implement and tune compared to OptiSLang, which involves multiple parameters for its various optimization algorithms. \textbf{2- Faster Convergence:} QDPSO demonstrates improved convergence speed due to its quantum-inspired search mechanism, allowing for quicker identification of optimal antenna designs compared to the more complex algorithms in OptiSLang. \textbf{3- Effective Exploration of Solution Space:} The quantum behaviors in QDPSO enhance the exploration capabilities within the solution space, enabling it to avoid local minima more effectively than traditional methods employed in OptiSLang. \textbf{4- Robustness in Handling Nonlinear Problems:} QDPSO has shown superior performance in optimizing complex and nonlinear antenna design problems, which can be challenging for some of the algorithms available in OptiSLang. \textbf{5- Direct Application to Electromagnetic Problems:} QDPSO is specifically tailored for electromagnetic applications, making it highly effective for antenna design tasks. In contrast, OptiSLang is a general-purpose optimization tool that may require additional adaptation for specific antenna-related challenges. \textbf{6- Lower Computational Cost:} Due to its efficient search mechanism and fewer parameters, QPSO achieves optimal solutions with lower computational resources compared to the potentially higher demands of OptiSLang's multi-faceted optimization strategies.
\end{comment}
\subsection{Antenna Optimization Process}
\begin{figure}
    \centering
    \includegraphics[width=0.9\linewidth]{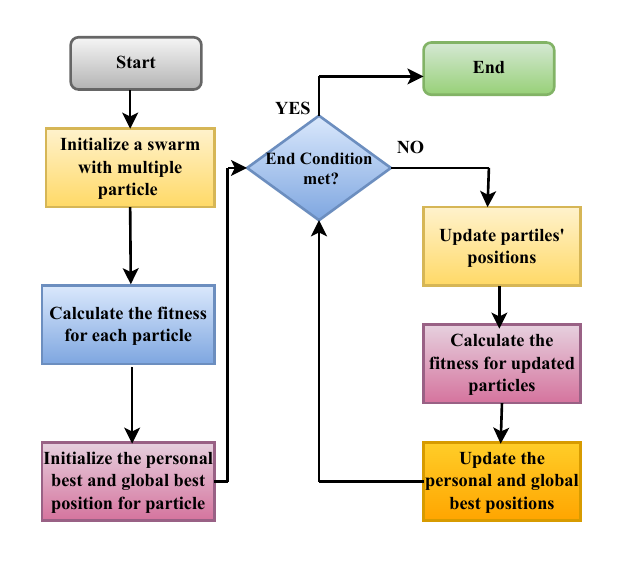}
    \caption{A Flowchart of the QDPSO}
    \label{flowqpso}
\end{figure}

In the context of slot antenna miniaturization using loops, the optimization focuses on the inner ($d_{\text{inner}} $) and outer ($ d_{\text{outer}} $) diameters of the loop, while other dimensions and substrate properties, listed in Table I, remain unchanged. As discussed in Section II, the goal is to minimize the resonance frequency of the loop-loaded slot topology.
\begin{comment}
In the context of slot antenna miniaturization utilizing loops, the optimization process is concentrated on the inner ($d_{\text{inner}} $) and outer ($ d_{\text{outer}} $) diameters of the loop, whereas all other dimensions and substrate properties, as outlined in Table I, remain unaltered. As outlined in Section II, the primary objective is to determine the optimal inner and outer diameters that can achieve the lowest possible resonance frequency for the slot with loops-loaded topology.
\end{comment}
The conventional process is difficult because it requires many trial-and-error simulations in ANSYS [43] to find the lowest resonance frequency, making the process very time-consuming. This optimization showcases the potential of automation to reduce engineering effort and shorten development cycles to a few hours. The flow chart of QDPSO is given in Fig. \ref{flowqpso}. The steps to get the best solutions are given as follows: \textbf{1- Initialization of Particles:} Each particle \( i \) is initialized with specific attributes. The position of the particle is determined by assigning the initial values for \( d_{\text{inner}} \) and \( d_{\text{outer}} \) randomly within their respective feasible ranges.
\begin{equation}
1.2 \, \text{mm} < d_{\text{inner}} < d_{\text{outer}} \leq 12 \, \text{mm}
\end{equation}
\begin{equation}
d_{\text{outer}} - d_{\text{inner}} > 0.8 \, \text{mm}
\end{equation}

%\text{1. Initialization of Particles}
%\begin{itemize}  
%Each particle \( i \) is initialized with specific attributes. The position of the particle is determined by assigning the initial values for \( d_{\text{inner}} \) and \( d_{\text{outer}} \) randomly within their respective feasible ranges.
%\begin{equation}
%1.2 \, \text{mm} < d_{\text{inner}} < d_{\text{outer}} \leq 12 \, \text{mm}
%\end{equation}
%\begin{equation}
%d_{\text{outer}} - d_{\text{inner}} > 0.8 \, \text{mm}
%\end{equation}
Specific geometric constraints are implemented to preserve the structural integrity and optimize the performance of the antenna. The inner diameter must exceed 1.20 mm, consistent with the slot width, to prevent merging with the slot, which could compromise structural stability. Similarly, the outer diameter is restricted to a maximum of 12.0 mm, as the slot length measures 30 mm and the combined length of the two loops is 24 mm. Surpassing this limit would induce overlap with the slot, increasing the overall antenna length and undermining the objective of resonance frequency reduction.
\begin{comment}
In this design, specific geometric constraints are imposed to ensure structural integrity and optimal performance of the antenna. First, the inner diameter is required to be greater than 1.20 mm, as the slot width is defined to be 1.20 mm. Maintaining this threshold prevents the inner diameter from merging with the slot, which could otherwise compromise the design’s structural stability and diminish its functional effectiveness. Similarly, an upper limit is established for the outer diameter, which must not exceed 12.0 mm. Given that the slot length is specified as 30 mm and the combined length occupied by the two loops is 24 mm, exceeding this boundary would result in an overlap with the slot. Such an overlap not only increases the overall antenna length but also contradicts the primary objective of achieving resonance frequency reduction.  
\end{comment}
Additionally, the difference between the outer and inner diameter must be no less than 0.80 mm to ensure feasibility for practical implementation. A smaller difference can cause fabrication issues and degrade antenna performance. Any particle position violating these limits is adjusted or resampled to stay within the defined parameter space, ensuring structural integrity and operational efficiency while meeting performance targets.
\begin{comment}
A smaller difference may lead to fabrication challenges and affect the antenna’s performance. To adhere to these design constraints, any particle position that breaches these limits is adjusted or resampled to remain within the defined parameter space. Enforcing these constraints ensures structural integrity and operational efficiency while achieving the targeted performance objectives.To maintain compliance with these design constraints, any particle position that violates these boundaries is either adjusted or re-sampled to remain feasible within the defined parameter space. By enforcing these constraints, the proposed design maintains structural robustness and operational efficiency while meeting the desired performance goals. 
\end{comment}
Then, the initial position of each particle is set as its personal best \( \mathbf{p}_i \). Among all particles, the one with the best fitness is chosen as the global best \( \mathbf{g} \).
\textbf{2- Evaluate Fitness: }The \text{fitness function} evaluates how well each particle’s current position meets the target specifications. The closer \( f_r \) is to the target frequency \( f_{\text{target}} \), the higher the fitness. Additional constraints, such as return loss, and efficiency, may further influence the evaluation.
%\end{itemize}
\textbf{3- Update Global and Personal Bests: }If a particle’s current position achieves a higher fitness than any previous position it held, this position becomes its new personal best \( \mathbf{p}_i \). Among all particles, the position with the highest fitness is updated as the new global best \( \mathbf{g} \).
\begin{comment}
  \item \text Personal Best: If a particle’s current position achieves a higher fitness than any previous position it held, this position becomes it is new personal best, \( \mathbf{p}_i \).
    \item \text Global Best: Among all particles, the position with the highest fitness is updated as the new global best \( \mathbf{g} \).
\end{comment}
\textbf{4- Quantum-Inspired Position Update: } Unlike classic PSO, QDPSO uses a quantum model for updating positions. Each particle’s new position is probabilistically determined to balance exploration and convergence is given by: 
\begin{equation}
\mathbf{x}_i^{(t+1)} = \mathbf{p}_i + \beta \cdot |\mathbf{p}_i - \mathbf{g}| \cdot \ln\left(\frac{1}{u}\right)
\end{equation}

where, the \(x_i^{(t+1)}\) represents the new position of the particle  \(i\) at time \(t+1\),  the \(p_i\) shows the personal best position of particle and \( \beta \) is a contraction-expansion coefficient. The \(g\) shows the global best position and variable \( u \) is a random number uniformly distributed between \( 0 \) and \( 1 \)  the  term \( |\mathbf{p}_i - \mathbf{g}| \) represents the Euclidean distance between the particle’s personal best and the global best, encouraging the particle to move closer to the globally optimal region.

\begin{comment}
    \item \( \beta \) is a contraction-expansion coefficient.
    \item \( u \) is a random number uniformly distributed between \( 0 \) and \( 1 \).
    \item The term \( |\mathbf{p}_i - \mathbf{g}| \) represents the Euclidean distance between the particle's personal best and the global best, encouraging the particle to move closer to the globally optimal region.
\end{comment}

%\end{itemize}

This probabilistic update makes the particles "quantum-behaved," allowing them to potentially explore new areas within the search space instead of being confined by velocity-based movement.
\textbf{5- Convergence Check: }The algorithm iterates through steps 2-5 until it meets a convergence criterion, which could be: The process continues for a set number of iterations. It stops if there is minimal improvement in the global best fitness over a set number of iterations. The process can also stop if a fitness value reaches a defined threshold.
\begin{comment}
    \item A set number of iterations,
    \item Minimal improvement in global best fitness over a set number of iterations,
    \item A fitness value reaching a defined threshold.
\end{comment}
%\end{itemize}
\textbf{6- Final Solution: }Once the algorithm converges, the global best position represents the optimal values for \( d_{\text{inner}} \) and \( d_{\text{outer}} \) that achieve the desired antenna performance. This solution balances the resonant frequency, return loss, bandwidth, and radiation efficiency.
\par
Each particle in QDPSO navigates the search space using a probabilistic, quantum-inspired position update. Particles adapt their positions based on personal and global best experiences, balancing exploration and exploitation to optimize the antenna design effectively. The search process effectively converges to the optimal inner ($d_{\text{inner}}$) and outer ($d_{\text{outer}}$) diameters, enhancing antenna performance by minimizing the resonant frequency while adhering to specified design criteria.
\subsection{Fitness function definition and evaluation}
%\section{Objective Function}
The goal is to get the minimum resonant frequency $f_r$ of the antenna for a given target frequency $f_{\text{target}}$  that is 2.27 GHz. As our aim is to get a minimum $f_r$ value, we have to get the maximum absolute difference between $f_r$ and $f_{\text{target}}$. In equation 4, a novel fitness function is designed for that purpose. 

\begin{equation}
\text{Minimize Fitness} = \frac{1}{1 + |f_r - f_\text{target}|}
\end{equation} \label{eq4}\\
The fitness function is designed to reward solutions that maximize the difference between the resonant frequency (\(f_r\)) and the target frequency $f_{\text{target}}$. The term \( |f_r - f_\text{target}| \) represents the absolute difference, and a larger absolute difference results in a lesser fitness value, encouraging the QDPSO algorithm to get a minimum $f_r$ value. Adding 1 to the denominator ensures that the fitness value remains finite and avoids division by zero when \( |f_r - f_\text{target}| = 0 \), where the fitness value reaches its maximum of 1. 
\\
Here, $f_r$ (the resonant frequency) is a function of the antenna’s dimensions and the effective dielectric constant $\epsilon_{\text{eff}}$, and is calculated as:

\begin{equation}
f_r = \frac{c}{2 \times (L_{SL} + d_{\text{outer}}) \times \sqrt{\epsilon_{\text{eff}}}}
\end{equation}

The effective dielectric constant $\epsilon_{\text{eff}}$ is approximated by:
\begin{equation}
\epsilon_{\text{eff}} = \frac{\epsilon_r + 1}{2} + \frac{\epsilon_r - 1}{2} \left( 1 + \frac{12 h}{W} \right)^{-0.5}
\end{equation}

\begin{table}[ht]
    \caption{Outer Diameter, Inner Diameter, Frequency, and Miniaturization Results}
    \centering
    \renewcommand{\arraystretch}{1.5} % Increase row height
    \begin{tabular}{c@{\hspace{10pt}}c@{\hspace{10pt}}c@{\hspace{10pt}}c}
        \hline
        \textbf{Outer Diameter} & \textbf{Inner Diameter} & \textbf{Frequency} & \textbf{Miniaturization } \\
        \hline
        11.9787 mm & 2.9498 mm & 1.5912  GHz & 29.90 \% \\ 
        11.6620 mm & 7.0070 mm & 1.5711  GHz & 30.79 \% \\  
        11.5439 mm & 4.9963 mm & 1.5411  GHz & 32.11 \% \\ 
        11.9711 mm & 3.2773 mm & 1.5110  GHz & 33.44 \% \\ 
        11.8729 mm & 5.2052 mm & 1.5110  GHz & 33.44 \% \\
        11.1002 mm & 2.2014 mm & 1.4810  GHz & 34.76 \%  \\ 
        11.9187 mm & 5.6182 mm & 1.4409  GHz & 36.52 \% \\ 
        11.8614 mm & 6.2441 mm & 1.4208  GHz & 37.41 \% \\ 
        \hline
    \end{tabular}
    \label{table:results}
\end{table}

\begin{figure}
    \centering
    \includegraphics[width=0.5\linewidth]{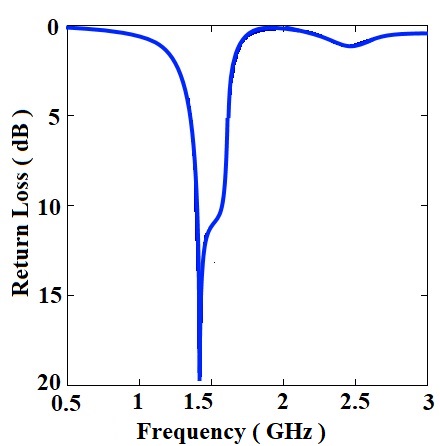}
    \caption{Return Loss for loop loaded slot antenna  with Outer Diameter of 11.8614 mm and Inner Diameter of 6.2441 mm }
    \label{fig:enter-label}
\end{figure}

\begin{figure}[h]
    \centering
    \includegraphics[width=0.9\linewidth]{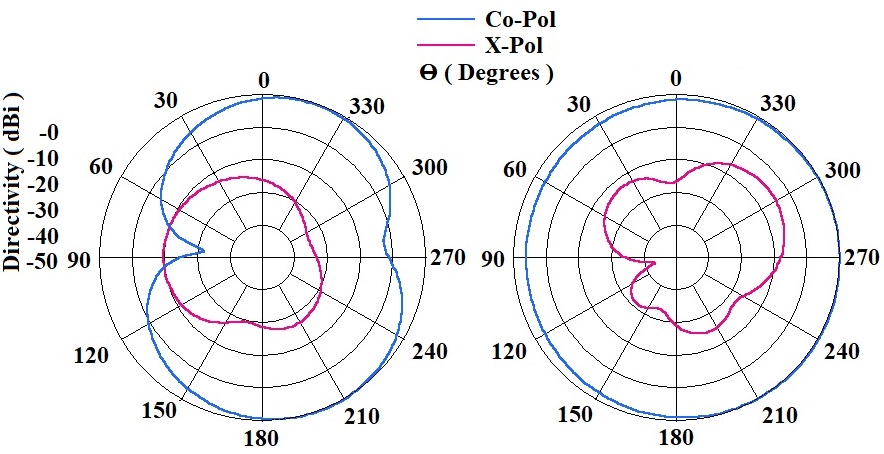}
    
    \vspace{1mm} % small space between image and labels
    \begin{tabular}{@{}c@{}}
        (a) E-Plane \hspace{2.8cm} (b) H-Plane
    \end{tabular}
    \caption{Radiation pattern at frequency 1.4208 GHz}
    \label{fig:eh-plane}
\end{figure}
\vspace{1mm}
\subsection{ Optimization of  slot antenna design using QDPSO and ANSYS }
The QDPSO algorithm generates potential solutions for the inner $ (d_{\text{inner}}) $ and outer ($d_{\text{outer}} $) diameters of the loop, which are identical at both ends of the slot. These solutions are then implemented in ANSYS to determine the corresponding resonance frequencies. The results obtained from simulations using ANSYS, along with frequency reduction data, are analyzed in comparison to the reference slot's frequency of 2.27 GHz, as detailed in Table II. This table delineates the responses for only 8 samples, although QDPSO is capable of generating a multitude of inner and outer diameter combinations, offering significant potential for optimal miniaturization. In comparison to Fig. 2 in the literature [7], the resonance frequency has been further reduced from 1.60 GHz to 1.4208 GHz, despite introducing less number of loops at either end of the slot.A comparative analysis proposes an alternative loop-loaded slot antenna design, improving miniaturization, time efficiency, and AI-CAD integration for rapid wireless devices development in a very short time. The return loss the loop loaded slot antenna, characterized by an outer diameter  ($d_{\text{outer}}$)  of 11.8614 mm and an inner diameter $ (d_{\text{inner}}) $  of 6.2441 mm, is depicted in Fig. 5, which demonstrates the maximum miniaturization achieved. The corresponding simulated ANSYS responses for the E-Plane and H-Plane are also shown in Fig. 6. Additionally, the QDPSO response achieved superior miniaturization compared to Fig. 8 in [5], demonstrating a 19.33\% reduction for the slot antenna. 
\begin{table}[h]
    \centering
    \caption{Constant Parameter Specifications for Data Set Creation}
    \begin{tabular}{@{}ll@{}}
        \toprule
        \textbf{Parameter} & \textbf{Specification} \\ 
        \midrule
        \textbf{Substrate Material Properties} & \\
        Relative Permittivity ($\varepsilon_r$) & 10.2 (Dielectric Constant) \\
        Thickness & 2.54 mm \\
        Loss Tangent ($\tan \delta$) & 0.0023 (Dielectric Losses) \\ 
        \midrule
        \textbf{Slot Design Parameters} & \\ 
        Position (X, Y, Z) & (-0.6, -15, 0) Centrally on ground \\ 
        Dimensions & Length: 30 mm, Width: 1.2 mm\\ 
        \midrule
        \textbf{Antenna Placement and Dimensions} & \\ 
        Position (X, Y, Z) & (-50, -50, 0) \\ 
        Dimensions & Length: 100 mm, Width: 100 mm\\ 
        \midrule
        \textbf{Loop Parameters} & \\ 
        First Loop Position (X, Y, Z) & (0, -9, 0) \\ 
        Second Loop Position (X, Y, Z) & (0, 9, 0) \\ 
        Loop Placement & Positioned at either end of the slot \\ 
        \midrule
        \textbf{Microstrip Feed Line} & \\ 
        Position (X, Y, Z) & (-3.6, 9, -2.54) \\ 
        Dimensions & Length: 53.6 mm, Width: 2.3 mm \\ 
        Microstrip feed Placement & Positioned at bottom surface \\ 
        \bottomrule
    \end{tabular}
\end{table}
\vspace{-3mm}
\section{Data Sets And Machine Learning Algorithms}
A primary challenge in applying machine learning to antenna design is the limited availability of datasets. While standard designs offer restricted data, bespoke antennas often necessitate the generation of proprietary datasets, which are crucial for developing precise and reliable models. In this study, the dataset for the proposed antenna was synthesized using ANSYS Electronics Desktop [43], comprising 936 data points. The design primarily emphasizes two critical parameters—the inner and outer diameters of the loop—which play a pivotal role in achieving optimal miniaturization, as depicted in Fig. 1. The dataset was generated by varying these two diameters and captures four key response variables: (1) resonance frequency, (2) return loss, (3) return loss depth, and (4) efficiency. All other dimensions were held constant, as detailed in Table III. Particular attention is given to the inner and outer diameters due to their significant influence on frequency reduction relative to the $\lambda$/2 slot resonance at 2.27 GHz. This focused investigation of the diameter dimensions is fundamental to enhancing both the miniaturization and overall performance of the antenna.
\begin{figure}[h]
    \centering
\includegraphics[width=1.05\linewidth]{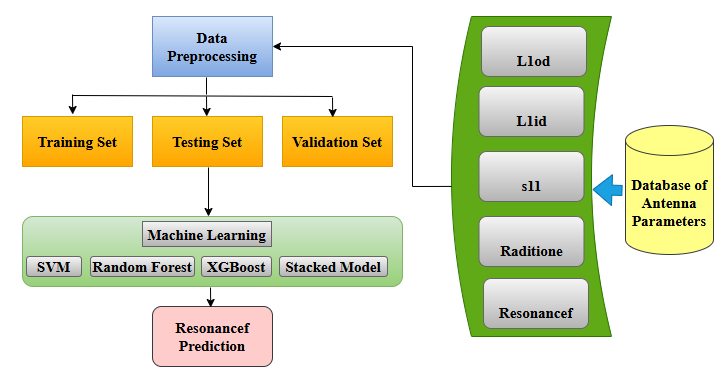}
    \caption{Steps to Predict the output}
    \label{flowml}
\end{figure}
%\vspace{1cm}
The subsequent analysis employs machine learning algorithms, such as SVM [44], Random Forest [45], XG Boost[46], and Stacked Model [47] to predict resonance frequency using 936 datasets.The algorithms used to predict the resonance frequency are detailed in the appendices A. The steps to calculate the predicted value of resonance frequency are given in Fig. \ref{flowml}.
%%%%%%%%%%%%%%%%%%%%%%%%%%%%%%%%%%%%%%

\section{ Results of Evaluation by Machine Learning framework}
\begin{figure}
    \centering
    \includegraphics[width=1\linewidth]{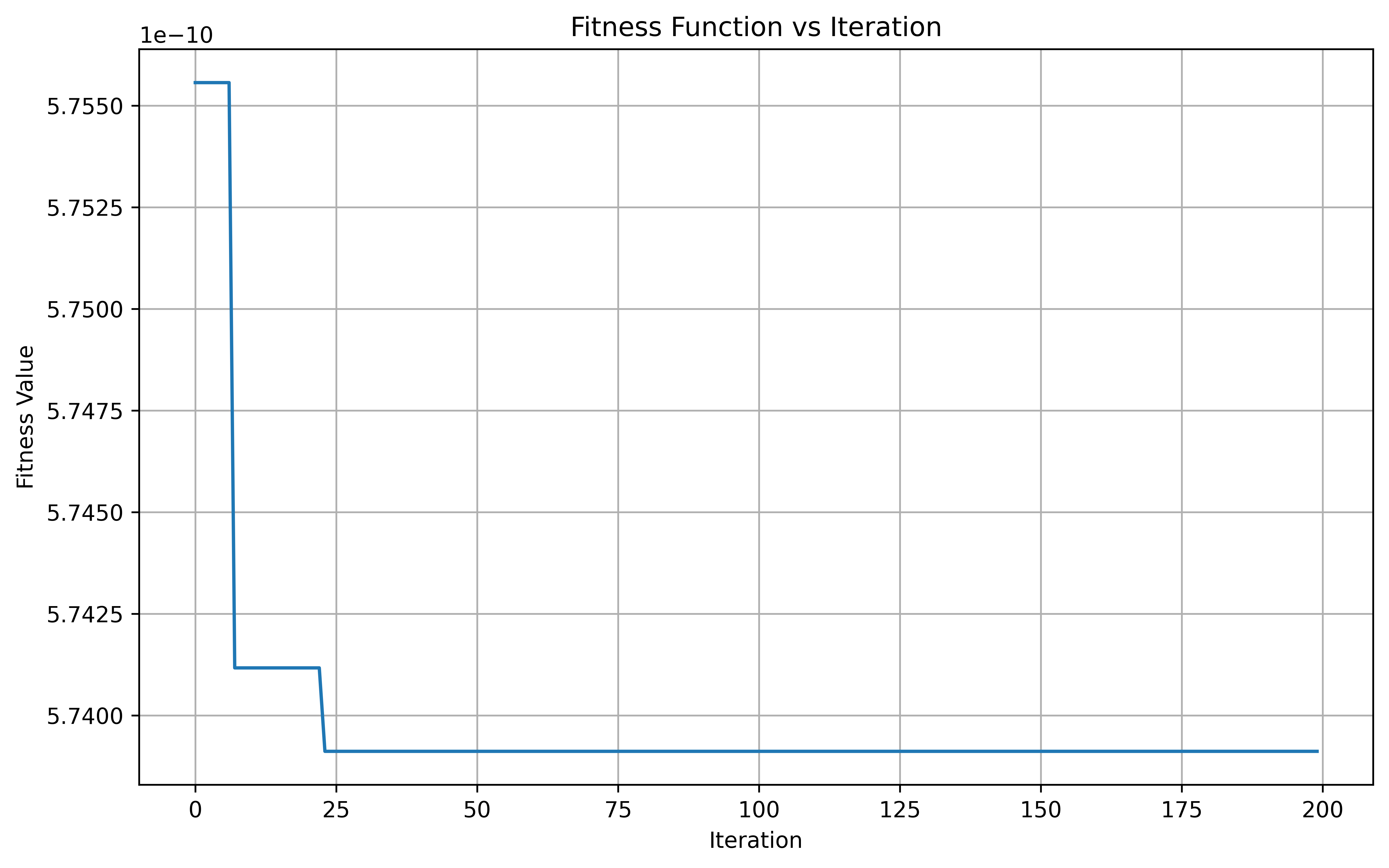}
    \caption{Fitness vs No. of Iterations}
    \label{fitness}
\end{figure}

Fig. \ref{fitness} illustrates the progression of a fitness function. The x-axis represents the number of iterations, while the y-axis shows the fitness value. The fitness value decreases as the absolute difference \(|f_r - f_\text{target}|\) increases.
In the initial phase (0--20 iterations), the fitness value improves rapidly, indicating that the algorithm effectively increases the difference \(|f_r - f_\text{target}|\). This suggests that the initial solutions were far from the expected result, however, the algorithm quickly identifies promising regions of the solution space. The fitness value stabilizes in the plateau phase (after approximately 20 iterations), implying that the algorithm has maximized the difference to a nearly constant value.
The data is split into 90\% training data, 5\% validation data, and 5\% testing data. The large training dataset allows models to capture comprehensive patterns, while the validation set aids in fine-tuning hyperparameters and preventing overfitting. The base models are trained on the Training set, generating predictions for the Training, Testing, and Validation sets. These predictions are then combined to create stacked datasets for each split: $S_{train}$, $S_{test}$, and $S_{val}$. A detailed explanation of different machine learning models is given in Appendices A, B, and C. The performance of all models are evaluated using various metrics such as Mean Absolute Error ( MAE ), Mean Squared Error 
 ( MSE ), Root Mean Squared Error( RMSE ), Coefficient of Determination ( R² ), Root Mean Squared Percentage Error ( RMSPE ), and Mean Absolute Percentage Error ( MAPE ) for the Training, Testing, and Validation sets. The use of MAE, MSE, RMSE, R², RMSPE, and MAPE is crucial in evaluating the performance of machine learning models as they provide various perspectives on accuracy, error distribution, and the model's ability to generalize. These metrics help assess the deviation between predicted and actual values, ensuring that models are evaluated comprehensively across different error types. The detailed  results are displayed in tables \ref{tab:performance_metrics_train}, \ref{tab:performance_metrics_test} and \ref{tab:performance_metrics_val}.
%%%%%%%%%%%%%%%%%%%%%

\begin{figure}
    \centering
    \includegraphics[width=1\linewidth]{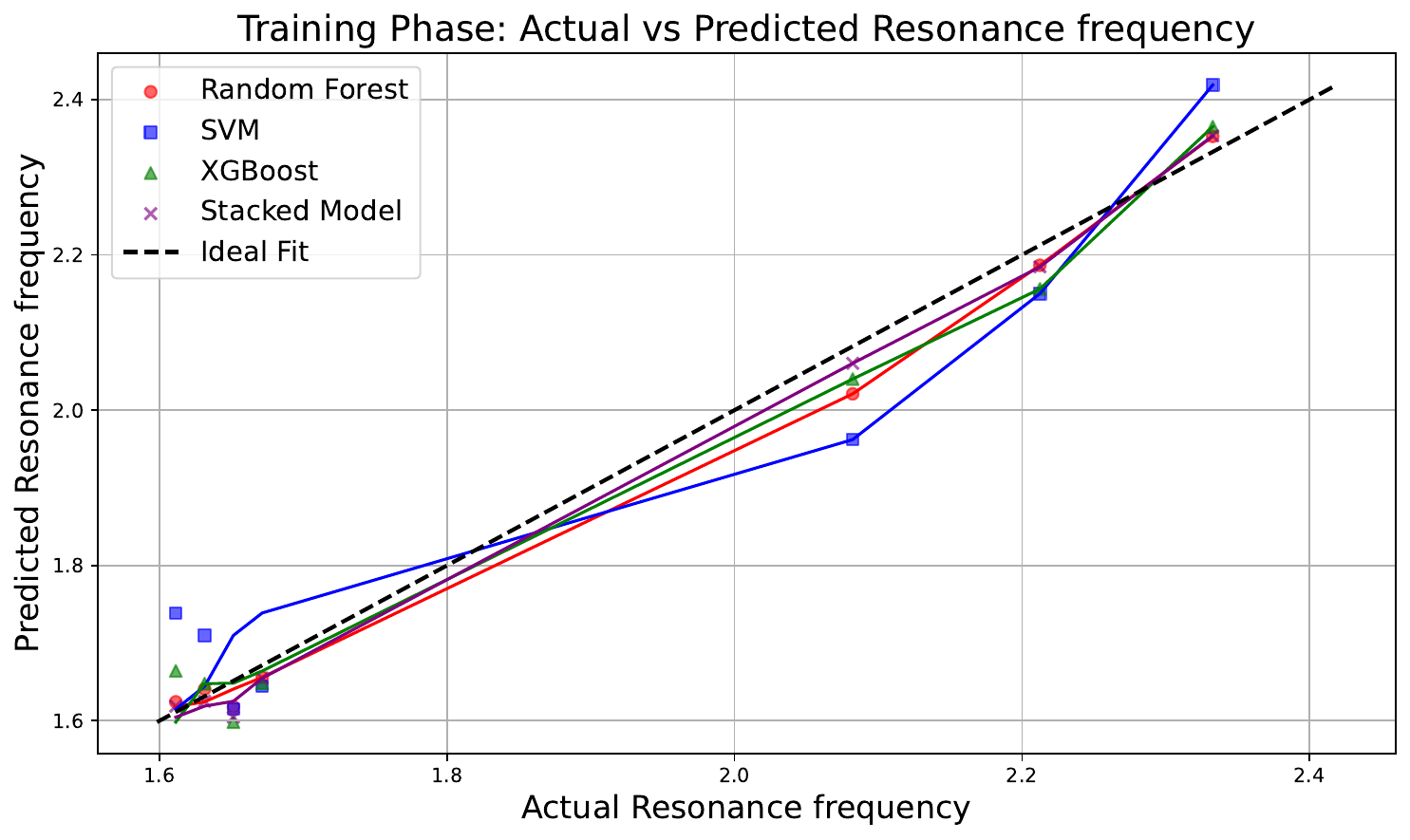}
    \caption{Scatter plots of the different models during the Training phase}
    \label{scatter3}
\end{figure}

\begin{table}[ht]
\caption{Performance Metrics of Models on Training Data}
\centering
\begin{tabular}{lcccccc}
\toprule
\textbf{Model} & \textbf{MAE} & \textbf{MSE} & \textbf{RMSE} & \textbf{R²} & \textbf{RMSPE} & \textbf{MAPE} \\
\midrule
Random \\Forest & 0.0411 & 0.0046 & 0.0679 & 0.9410 & 4.7731 & 2.4355 \\ \\
SVM & 0.1020 & 0.0264 & 0.1625 & 0.6616 & 11.3896 & 6.0921 \\ \\
XGBoost & 0.0353 & 0.0026 & 0.0505 & 0.9673 & 3.2191 & 2.0333 \\ \\
Stacked\\ Model & 0.0276 & 0.0014 & 0.0369 & 0.9825 & 2.1828 & 1.5492 \\ \\
\bottomrule
\end{tabular}
\label{tab:performance_metrics_train}
\end{table}
\vspace{3mm}
\begin{table}
\caption{Performance Metrics of Models on Testing Data}
\centering
\begin{tabular}{lcccccc}
\toprule
\textbf{Model} & \textbf{MAE} & \textbf{MSE} & \textbf{RMSE} & \textbf{R²} & \textbf{RMSPE} & \textbf{MAPE} \\
\midrule
Random \\Forest & 0.1115 & 0.0236 & 0.1536 & 0.6938 & 9.2578 & 6.2138 \\ \\
SVM & 0.0990 & 0.0190 & 0.1378 & 0.7537 & 8.0458 & 5.4768 \\ \\
XGBoost & 0.1196 & 0.0288 & 0.1697 & 0.6263 & 10.1062 & 6.5767 \\ \\
Stacked \\ Model & 0.1232 & 0.0303 & 0.1740 & 0.6075 & 10.4648 & 6.8221 \\
\bottomrule
\end{tabular}
\label{tab:performance_metrics_test}
\end{table}
\vspace{3mm}
\begin{table}[ht]
\caption{Performance Metrics of Models on Validation Data}
\centering
\begin{tabular}{lcccccc}
\toprule
\textbf{Model} & \textbf{MAE} & \textbf{MSE} & \textbf{RMSE} & \textbf{R²} & \textbf{RMSPE} & \textbf{MAPE} \\
\midrule
Random \\ Forest & 0.0933 & 0.0170 & 0.1303 & 0.6110 & 8.0510 & 5.3656 \\ \\
SVM & 0.0858 & 0.0122 & 0.1106 & 0.7197 & 6.8617 & 4.9457 \\ \\
XGBoost & 0.0931 & 0.0190 & 0.1380 & 0.5638 & 8.8563 & 5.4557 \\ \\
Stacked \\ Model & 0.0953 & 0.0219 & 0.1480 & 0.4982 & 9.3922 & 5.5876 \\
\bottomrule
\end{tabular}
\label{tab:performance_metrics_val}
\end{table}
In the Fig.\ref{scatter3} Based on the R² values on the training data, the stacked model demonstrates the best performance with an R² of 0.9825, indicating that it explains approximately 98.25\% of the variance in the training data. This highlights the effectiveness of combining predictions from multiple models, as the stacked model leverages their strengths while minimizing individual weaknesses. XGBoost follows closely with an R² of 0.9673, showing its robustness and strong capability to model complex patterns. Random Forest, with an R² of 0.9410, performs well but is slightly less than XGBoost and the stacked model. In contrast, SVM exhibits the weakest performance on the training data with an R² of 0.6616, and its higher RMSE, RMSPE, and MAPE values indicate it struggles to capture the non-linear relationships in the data. Overall, on the training data, the stacked model and XGBoost emerge as the most effective approaches, with Random Forest being a reliable alternative.
\par
\begin{figure}
    \centering
    \includegraphics[width=1\linewidth]{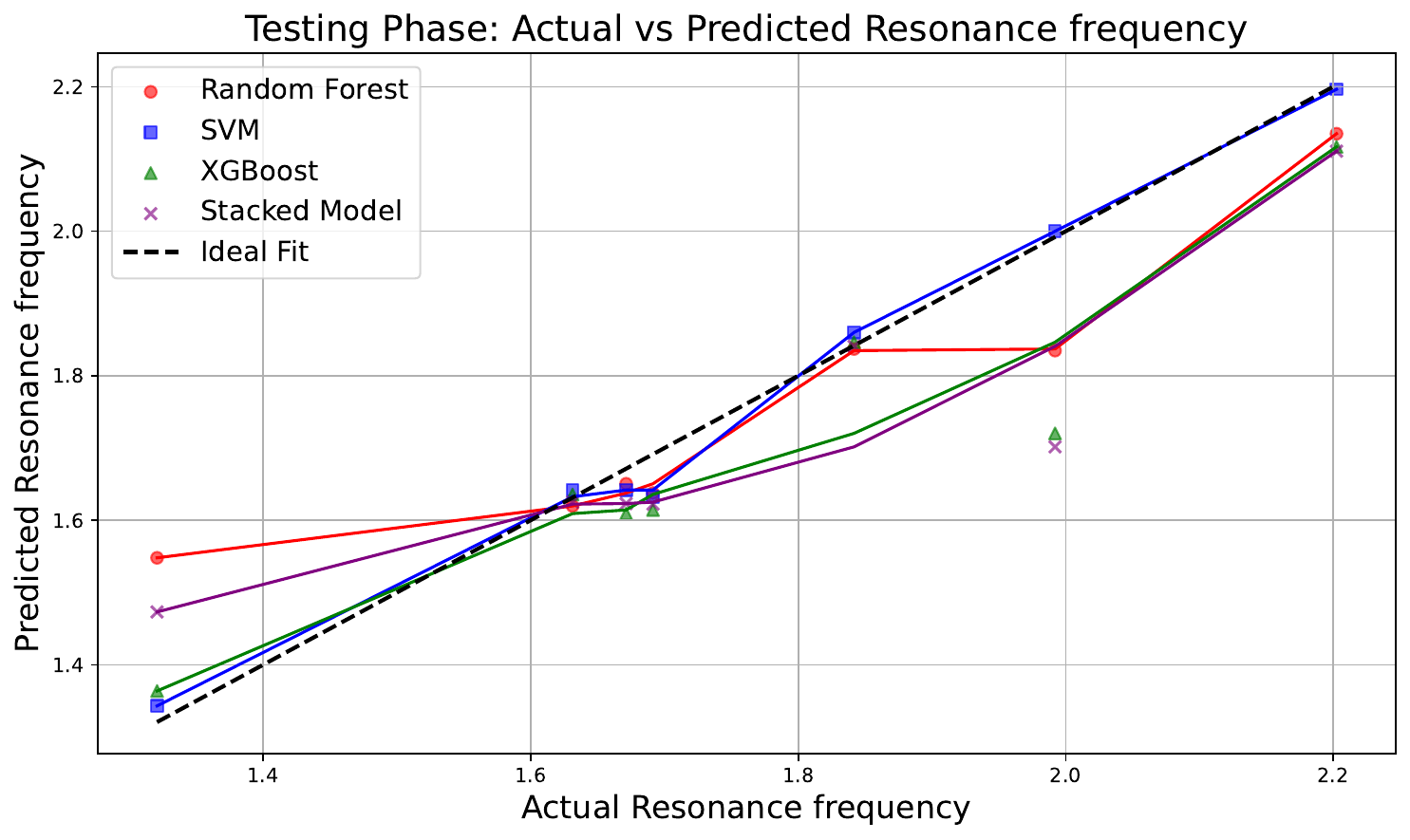}
    \caption{Scatter plots of the different models during the Testing phase}
    \label{scatter2}
\end{figure}

In Fig.\ref{scatter2} on the validation data, the performance of the models based on R² values reveals distinct strengths and weaknesses. SVM demonstrates the best performance with an R² of 0.7197, indicating that it explains approximately 71.97\% of the variance in the validation data. This highlights its capability to generalize well and effectively model the relationships in unseen data. Random Forest follows with an R² of 0.6110, showing moderate predictive ability, though it captures less variance compared to SVM. XGBoost, with an R² of 0.5638, performs slightly worse than Random Forest, suggesting its limited ability to handle the complexity of the validation data in this scenario. The stacked model, with the lowest R² of 0.4982, struggles the most on the validation data, explaining only 49.82\% of the variance. This indicates that while the stacked model performed exceptionally well on the training data, it may be prone to overfitting, reducing its effectiveness on unseen data. Overall, SVM emerges as the most robust model on the validation data, followed by Random Forest and XGBoost, while the stacked model falls short in terms of generalization.
\par
\begin{figure}
    \centering
    \includegraphics[width=1\linewidth]{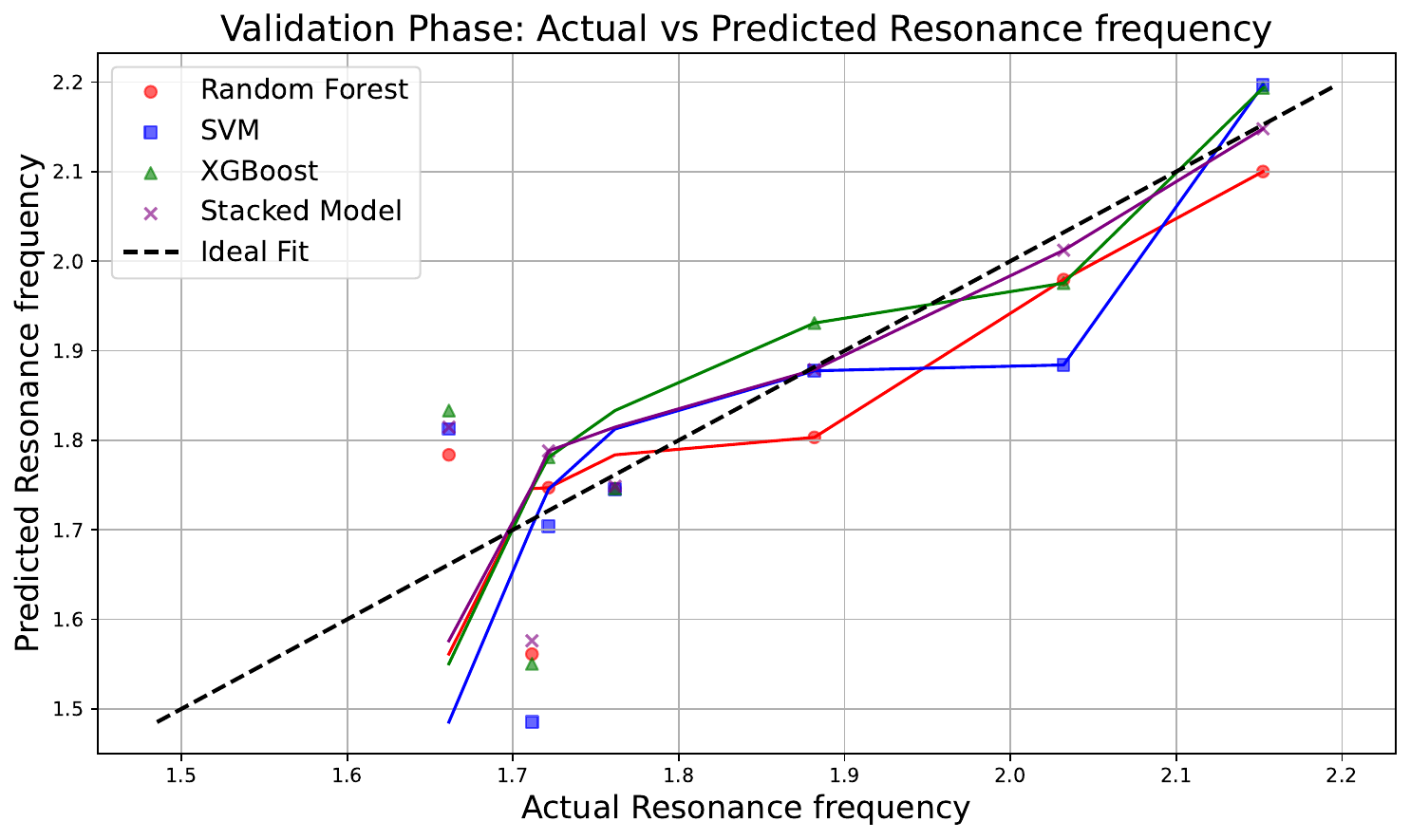}
    \caption{Scatter plots of the different models during the validation phase}
    \label{scatter1}
\end{figure}
\par
\vspace{1mm}
In Fig.\ref{scatter1} the SVM model outperforms the others with the highest R² value of 0.7197, explaining 71.97\% of the variance in the actual resonance frequency values. This suggests that the SVM model provides the best fit, with the scatter plot showing tightly clustered points along the ideal fit line, indicating a strong correlation between predicted and actual values. The Random Forest model follows with an R² of 0.6110, explaining 61.1\% of the variance, and its scatter plot shows moderate clustering around the diagonal line with some spread. XGBoost performs the worst with an R² of 0.5638, indicating that it explains only 56.38\% of the variance, and the scatter plot for this model would show a greater spread of points, reflecting a weaker correlation between predictions and actual values. Surprisingly, the stacked model, which combines the predictions of the three models, performs the worst overall with an R² of 0.4982, indicating that it explains less than half of the variance. The scatter plot for the stacked model would show more dispersed points, reflecting its poorer fit compared to the individual models.
%%%%%%%%%%%%%%%%%5%%%
\begin{table}[ht]
    \centering
    \renewcommand{\arraystretch}{1.5} % Increase row height
    \caption{Predicted and Actual Frequency}
    \label{table:antenna_predictions}
    \begin{tabular}{|c|c|c c|}
        \hline
        \textbf{Outer Diameter} & \textbf{Inner Diameter} & 
        \multicolumn{2}{c|}{\textbf{Frequency (GHz)}} \\ 
        \hline
        \textbf{} & \textbf{} & \textbf{Predicted} & \textbf{ANSYS} \\ 
        \hline
        11.9787 mm & 2.9498 mm & 
        \begin{tabular}[c]{@{}c@{}}RF: 1.6290 \\ SVM: 1.6135 \\ XGB: 1.6327 \\ SM: 1.6371\end{tabular} & 
        \textbf{1.5912} \\ 
        \hline
        11.6620 mm & 7.0070 mm & 
        \begin{tabular}[c]{@{}c@{}}RF: 1.6309 \\ SVM: 1.6904 \\ XGB: 1.6748 \\ SM: 1.6457\end{tabular} & 
        \textbf{1.5711} \\ 
        \hline
        11.5439 mm & 4.9963 mm & 
        \begin{tabular}[c]{@{}c@{}}RF: 1.6209 \\ SVM: 1.6687 \\ XGB: 1.6194 \\ SM: 1.6053\end{tabular} & 
        \textbf{1.5411} \\ 
        \hline
        11.9711 mm & 3.2773 mm & 
        \begin{tabular}[c]{@{}c@{}}RF: 1.5984 \\ SVM: 1.5980 \\ XGB: 1.5756 \\ SM: 1.5815\end{tabular} & 
        \textbf{1.5110} \\ 
        \hline
        11.8729 mm & 5.2052 mm & 
        \begin{tabular}[c]{@{}c@{}}RF: 1.6500 \\ SVM: 1.6664 \\ XGB: 1.6701 \\ SM: 1.6607\end{tabular} & 
        \textbf{1.5110} \\ 
        \hline
        11.1002 mm & 2.2014 mm & 
        \begin{tabular}[c]{@{}c@{}}RF: 1.6237 \\ SVM: 1.5690 \\ XGB: 1.6398 \\ SM: 1.6531\end{tabular} & 
        \textbf{1.4810} \\ 
        \hline
        11.9187 mm & 5.6182 mm & 
        \begin{tabular}[c]{@{}c@{}}RF: 1.6966 \\ SVM: 1.6441 \\ XGB: 1.6880 \\ SM: 1.7080\end{tabular} & 
        \textbf{1.4409} \\ 
        \hline
        11.8614 mm & 6.2441 mm & 
        \begin{tabular}[c]{@{}c@{}}RF: 1.5529 \\ SVM: 1.6576 \\ XGB: 1.6751 \\ SM: 1.6109\end{tabular} & 
        \textbf{1.4208} \\ 
        \hline
    \end{tabular}
\end{table}
%%%%%%%%%%%%%%%%%%%%%%%%%
Table \ref{table:antenna_predictions} presents a comparative analysis of predicted resonance frequencies using four machine learning models, Random Forest (RF), Support Vector Machine (SVM), XGBoost (XGB), and Stacked Model (SM), against the actual frequencies obtained from ANSYS simulations. The predictions were evaluated for different combinations of outer and inner diameters of a circular structure. Across all cases, the predicted frequencies closely approximate the ANSYS results, with minor deviations observed. Notably, RF and SM generally exhibit smaller deviations from the actual frequencies, indicating robust predictive performance. However, SVM and XGB occasionally show larger discrepancies, particularly in cases with complex diameter configurations, such as for higher inner diameters. This analysis highlights the effectiveness of machine learning models, especially ensemble approaches like RF and XGB, in approximating high-fidelity simulations with a balance of accuracy and computational efficiency.\par
In this context, the SVM (Support Vector Machine) exhibited strong generalization, demonstrating its capability to handle non-linear relationships through kernel functions. XGBoost performed well, but its slight sensitivity to hyperparameters and potential overfitting on the validation data was noted. Random Forest delivered competitive results, with its ensemble learning ensuring stability but slightly underperforming compared to SVM due to less effective handling of complex non-linearities. The Stacked Model, combining predictions from base models, showed excellent training performance but slightly overfitted the data, as indicated by its lower validation performance. Thus, while training data optimizes model performance, validation ensures generalization and testing data verifies real-world applicability.
\vspace{-4mm}
\section{Rapid Wireless Device Production Through Automated Antenna Design}
The rapid advancement of modern technology is driving a shift toward autonomous systems, including self-driving cars and automated drones. Similarly, Electronic Design Automation (EDA) [48] has transformed semiconductor development by automating complex design tasks, enabling the creation of highly sophisticated systems with billions of components. This highlights the need for similar automation in wireless consumer electronics, particularly in antenna design, to enhance efficiency and performance. An advanced automated antenna design system can be developed to allow users to specify performance requirements, upon which the system autonomously generates optimized design parameters along with precise fabrication instructions. In this study, the QDPSO optimization algorithm efficiently determined a representative set of loop dimensions within 11.53 seconds. Thereafter, the machine learning model predicted the corresponding resonance frequency in 0.75 seconds. The comprehensive design validation, performed in ANSYS and informed by the outputs of both the QDPSO and machine learning model, required 12 minutes 13.16 seconds prior to the commencement of antenna fabrication.The complete design finalization process required approximately 12.42 minutes, in contrast to the Parallel Surrogate-Assisted Differential Evolution Algorithm ( PSADEA ) [49], which demanded nearly 50 hours to achieve a comparable level of antenna design optimization.The work was performed on DESKTOP-8AAIDPT, equipped with an Intel Core i5-8500 (6 cores, 3.00 GHz), 16 GB RAM, and 64-bit Windows 10 Pro (Version 22H2) on an x64-based system.These prototypes enable rapid deployment in consumer and industrial applications, accelerating market entry. Conversely, in the absence of dimension predictions from QDPSO and frequency estimations from the ML model, the ANSYS design process becomes dependent on trial-and-error, resulting in an inherently unpredictable timeframe. While optimal results may occasionally be achieved quickly, the process often extends beyond a month, with experts typically requiring at least seven days, depending on design complexity. It can be claimed that automation in RF and microwave design boosts efficiency, accuracy, and integration, streamlining processes for faster, reliable production.
\vspace{-4mm}
\section{Conclusion}
This research introduces an advanced framework for antenna miniaturization by combining Quantum-Behaved Dynamic Particle Swarm Optimization (QDPSO) and Machine Learning (ML) techniques. The approach addresses challenges in automated wireless system design by optimizing loop dimensions through QDPSO, achieving a 12.7 percent reduction in resonance frequency (1.4208 GHz vs. conventional 1.60 GHz) with optimization completed in 11.53 seconds. ML models trained on 936 ANSYS HFSS simulations predict resonance frequencies in 0.75 seconds, with the stacked ensemble model showing high training accuracy (R² = 0.9825) and SVM excelling in validation generalization (R² = 0.7197). The entire design cycle—optimization, prediction, and validation—is executed in 12.42 minutes on standard hardware, representing a 240× speed improvement over the PSADEA benchmark (50 hours) and eliminating traditional trial-and-error methods that often span over a week. This efficiency enables rapid deployment of compact antennas for IoT, 5G, and industrial applications while maintaining performance metrics like bandwidth and gain. Future directions include integrating multi-physics simulations for holistic validation, developing cloud-based AI platforms for automated geometry generation, incorporating sustainable materials, and addressing multi-band or massive MIMO requirements using metamaterials and phased array optimizations. By bridging AI-driven optimization with CAD validation, the framework establishes a scalable blueprint for next-generation RF systems, reducing engineering workloads and accelerating prototype development. Its adaptability to diverse frequencies and form factors positions it as a critical tool for autonomous drones and smart city infrastructure, enhancing both design agility and technological innovation.

%This research presents a transformative approach to antenna miniaturization by integrating Quantum-Behaved Dynamic Particle Swarm Optimization (QDPSO) and machine learning techniques, addressing critical challenges in automated wireless system design. By optimizing loop dimensions using QDPSO, the study achieves a 12.7 percent resonance frequency reduction to 1.4208 GHz compared to conventional designs at 1.60 GHz, with optimization completed in just 11.53 seconds. Machine learning models predict resonance frequencies with inference times of 0.75 seconds, leveraging 936 ANSYS HFSS simulations to establish a robust data-driven framework. The stacked ensemble model demonstrates superior training accuracy (R² = 0.9825), while SVM outperforms in generalization on validation data (

% if have a single appendix:(R² = 0.9825)
% or



\ifCLASSOPTIONcaptionsoff
  \newpage
\fi

%\end{comment}



%\bibliographystyle{IEEEtran}

\begin{IEEEbiography}{Khan Masood Parvez} (Student Member, IEEE)was born in Birbhum, India, in 1993. He received his B.E., M.E., and Ph.D. degrees in Electronics and Communication Engineering from Aliah University, Kolkata, India, in 2015, 2017, and 2025, respectively. His current research interests include slot antennas, antenna miniaturization, cross-polarization improvement, electrically small antennas, and machine learning.
\end{IEEEbiography}
\begin{IEEEbiography}{Sk Md Abidar Rahaman} received the B.Tech, M.Tech, and Ph.D degrees in Computer Science and Engineering from Aliah University in the years 2013, 2015, and 2025 respectively. His research interests include artificial intelligence, machine learning, and related applications. Dr. Rahaman has published several papers in reputable conferences, book chapters, and journals. He is currently working as a faculty member of Tarakeswar Degree College, affiliated with the University of Burdwan, in the department of computer science.
\end{IEEEbiography}

\begin{IEEEbiography}{Ali Shiri Sichani} (Member, IEEE) received the Ph.D. degree in electrical engineering from the University of South Florida, Tampa, FL, USA, in 2021.He is currently an Assistant Teaching Professor with the Electrical Engineering and Computer Science Department, University of Missouri, Columbia, MO, USA. His research interests include neuromorphic computing and VLSI, as well as hardware design for AI applications.
\end{IEEEbiography}



%\vfill


\end{document}